# A Survey of Classification Techniques in the Area of Big Data.

[1]PrafulKoturwar, [2]SheetalGirase, [3]Debajyoti Mukhopadhyay

[1]Reseach Scholar, Department of Information Technology

[2]Assistance Professor, Department of Information Technology

[3]Head, Department of Information Technology

Maharashtra Institute of Technology Pune 411038, India

prafulkoturwar@gmail.com, girase.sheetal@gmail.com, debajyoti.mukhopadhyay@gmail.com

**A B S T R A C T**

Big Data concern large-volume, growing data sets that are complex and have multiple autonomous sources. Earlier technologies were not able to handle storage and processing of huge data thus Big Data concept comes into existence. This is a tedious job for users to identify accurate data from huge unstructured data. So, there should be some mechanism which classify unstructured data into organized form which helps user to easily access required data. Classification techniques over big transactional database provide required data to the users from large datasets more simple way. There are two main classification techniques, supervised and unsupervised. In this paper we focused on to study of different supervised classification techniques. Further this paper shows application of each technique and their advantages and limitations.

Index Terms: Big Data, Supervised Classification, Decision Tree, Support Vector Machine

## I. INTRODUCTION

Big Data is unstructured data that exceeds the processing complexity of conventional database systems. The data is too big, moves too fast, or doesn't fit the rule restricting behavior of our database architectures. This information comes from multiple, distinct, independent sources with complex and evolving relationships in a Big Data which is keep on growing day by day. There are three main challenges in Big Data which are data accessing and arithmetic computing procedures, semantics and domain knowledge for different Big Data applications and the difficulties raised by Big Data volumes, distributed data distribution and by complex and dynamic characteristics. Big data framework is divided into three tiers as shown in figure 1, to handle the above challenges [1].

Tier I which is data accessing and computing focus on data accessing and arithmetic computing procedures. Because large amount of information are stored at different locations which are growing rapidly day by day, hence for computing distributed large-scale of information we have to consider effective computing platform like Hadoop.

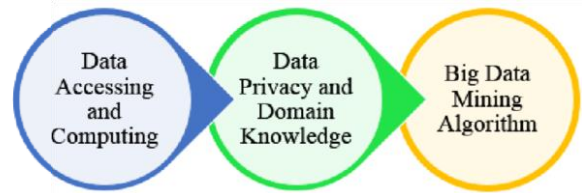

**Figure 1. Three Tiers of Big Data**

Data privacy and domain knowledge is the Tier II which focuses on semantics and domain knowledge for different Big Data applications [9]. In social network, users are linked with each other that shares their knowledge which are represented by user communities, leaders in each group and social influence modelling and so on, therefore for understanding their semantics and application knowledge is important for both low-level data access and for high-level mining algorithm designs.

Tier III which is Big Data mining algorithm focus on difficulties raised by Big Data volumes, distributed data distribution and by complex and dynamic characteristics. There are three stages in Tier III, as shown in Figure 2, (a) Sparse, heterogeneous, uncertain, incomplete and multisource data are pre-processed by data fusion techniques; (b) Complex and dynamic data are mined after pre-processing; (c) The global knowledge obtained by local learning and model fusion is tested and relevant information is feedback to the pre-processing stage [1].

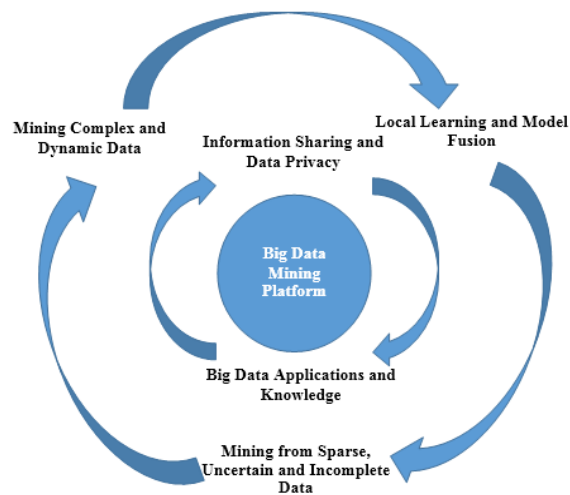

**Figure 2: Big Data Processing Framework**

Big Data is now part of every sector and function of the global economy [7]. Big Data a collection of datasets is so large and complex that is beyond the ability of typical database software tools to capture, store, manage and process the data within a tolerable elapsed time. A typical domain like stock market data are constantly generating a large quantity of information such as bids, buys and puts, in every single seconds [2]. This information impact on different factors such as domestic and international news, government reports and natural disasters and so on, hence it is nearly impossible to have required and appropriate information to user over such a complex and voluminous data so it is crucial that such a data should be classified appropriately and presented to the user for his convenience and ease of access.

Classification technique is used to solve the above challenges which classify the big data according to the format of the data that must be processed, the type of analysis to be applied, the processing techniques at work, and the data sources for the data that the target system is required to acquire, load, process, analyze and store [4]. Many classification techniques are used based on applications selected. Before actual classification begins, required information is extracted from large amount of data and then classification is done. There are two main classification techniques, supervised and unsupervised. Supervised classification techniques as shown in Figure 3, are also known as predictive or directed classification. In this method set of possible class is known in advanced. Unsupervised classification techniques are also known as descriptive or undirected. In this method set of possible class is unknown, after classification we can assign name to that class.

In supervised classification Decision Tree (DT) and Support Vector Machine (SVM) are well known classifier and used widely. Decision Tree is a hierarchical model that recursively does the separation of the input space into class regions. It consists of decision nodes and leaves. Learning algorithm for the Decision Tree is greedy, it finds the best attribute to split the data. Repeat this until it cannot be separated any more. The main aim of DT is to find out the smallest tree that would make the data after split as pure as possible. Support Vector Machine is a supervised method that analyzes data and recognizes patterns which is used for classification. Given a training set and the data needs to be classified into two classes, a SVM classifier builds a model that assigns the data into one of the categories. Extraction of huge training set is modelled as a multi-dimensional classification problem with one class for each action and its aim is to assign a class label to a given action or activity.

Rest of this paper is organized as follows: In Section 2 briefly describe the Overview of Classification over Big Data, Section 3 does Comparative study. Conclusion and Future scope is discussed in the last section.

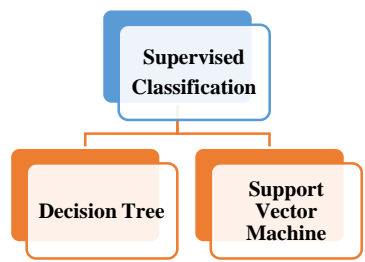

**Figure 3: Supervised Classification Techniques**

## 2. OVERVIEW OF CLASSIFICATION

Classification is one of the data mining technique that classifies unstructured data into the structured class and groups and it helps to user for knowledge discovery and future plan [3]. Classification provides intelligent decision making. There are two phases in classification, first is learning process phase in which a huge training data sets are supplied and analysis takes place then rules and patterns are created. Then the execution of second phase start that is evaluation or test of data sets and archives the accuracy of a classification patterns. This section briefly describes the supervised classification methods such as Decision Tree and Support Vector Machine.

## 2.1 SUPERVISED METHODS

Problems which involve classification are considered to be instances of a branch of machine learning called as "supervised learning" [5]. In this, the machine is given a "training set" of correctly classified instances of data in the first stage, and then the algorithm devised from this "learning" is used for the next stage of prediction. The converse of this is "unsupervised learning", which involves classifying data into categories based on some similarity of input parameters in the data.

Supervised Data Mining techniques are appropriate when we have a specific target value, so we can predict about our data [8]. The targets can have two or more possible outcomes, or even be a continuous numeric value. To use these methods, we ideally have a subset of training data (observations and measurement) sets for which this target value is already known. Training data includes both the input and desired results.

New data is classified based on the training set. The input data also called the training set which consists of multiple attributes or features. Each record is tagged with a class label. For some examples the correct results (target) are known and are given in input to the model during the learning process. The construction of a proper training validation and test set is crucial. These methods are usually fast and accurate.

The objective of classification is to analyze huge data and to develop an accurate description or model for each organized class using the feature present in the data. We use that training data to build a model of what a typical data set looks like when it has one of the various target values. We then apply that model to data for which that target value is currently unknown. The algorithm identifies the "new" data points that match the model of each target value. This model is used to classify test data for which the class descriptions are not known.

**1. Decision Tree (DT)**

Decision Tree is ideal to use as the filter to handle the large amount of data. DT is a basic way of classification can have satisfactory efficiency and accuracy of those datasets. Decision Tree algorithm is good at tuning between precision which can be trained very fast and provide sound results on those classification data [2].

Big Data are now rapidly expanding in all domains with the fast development of networking and increase in the data storage and collection capacity. The instances are divided into a set of discrete valued set of properties, known as various features of the data. For example, classifying a received email as "spam" or "not spam" could be based on analyzing characteristics of the email such as origin IP address, the number of emails received from the same origin, the subject line, the email address itself, the content of the body of the email, etc. All these features will contribute to a final value which will allow the algorithm to classify the email. It is logical that the more number of examples of spam and non-spam emails the Machine Learning system goes through, the better will be its prediction for the next unknown email.

Decision Tree learning is reasonably fast and accurate. The approach is to learn on large data sets is to parallelize the process of learning by utilizing Decision Trees. It is straightforward to reduce a Decision Tree to rules. The strategy follow here is to break a large data set into n partitions then learn a DT on each of the n partitions in parallel. A DT become bigger on each of n processors independently. After that they must be combined in such a way that the Decision Tree remains individual tree, for this approach Decision Tree can used Meta-learning. Meta-learning is the process by which learners become aware of and increasingly in control of habits of perception, inquiry, learning, and growth.

Now other aspect of creating final DT is pruning the tree which removes the nodes that do not provides accuracy in classification results in reduced size tree. Pruning is likely to be very important for large training set which will produce large trees. There are a number of methods to prune a Decision Tree. In C4.5 an approach called pessimistic pruning is quite fast and has been shown to provide trees that perform adequately.

Big Data challenges are growing day by day, traditional Decision Tree algorithms have multiple limitations. First, building a Decision Tree is a very time consuming when the available dataset is extremely huge.

Second, although parallel computing clusters can be leveraged in Decision Tree based classification algorithms, the strategy of data distribution should be optimized so that required data for building one node is localized and meanwhile the communication cost is minimized.

To overcome these limitations, distributed C4.5 algorithm with MapReduce computing model is used. When available dataset is extremely huge then C4.5 algorithm performs well in short time and it is robust in nature as well as simple to understand [10].

## 2. Support Vector Machine (SVM)

Machine Learning has ability to enable the computer to learn that uses algorithm and techniques which perform different tasks and activities to provide efficient learning. Our main problem is that how can we represent complex data and how to exclude bogus data. Support Vector Machine is a Machine Learning tool used for classification that is based on Supervised Learning which classifies points to one of two disjoint half-spaces. It uses nonlinear mapping to convert the original data into higher dimension. Its objective is to construct a function which will correctly predict the class to which the new point belongs and the old points belong. In the era of Big Data, the main reason behind maximum margin or separation because if we use a decision boundary to classify, it may end up nearer to one set of datasets compared to others. This happens only if data is structured or linear but mostly we find data is unstructured/nonlinear and dataset is inseparable then SVM kernels are used.

Traditional Classification approaches perform weakly when working directly because of huge amount of data but Support Vector Machine can avoid the problems of representing this much data. Support Vector Machine is the most promising technique and approach as compared to others classification approaches. Support Vector Machine balance proper and accurate huge amount of data and compromise between classifier complexity and error can be controlled explicitly. Another benefit of SVMs is that one can design and use a SVM kernel for a particular problem that could be applied directly to the data without the need for a feature extraction process. It is particularly important problems, where huge amount of structured data is lost by the feature extraction process.

Support Vector Machine (SVM) is the classification technique which used to process on large training data. The Big and complex data can be left to the SVM since the result of SVM will be greatly influenced when there is too much noise in the datasets. SVM provides with an optimized algorithm to solve the problem of over fitting. SVM is an effective classification model is useful to handle those complex data. SVM can make use of certain kernels to reveal efficiently in quantum form the largest eigenvalues and corresponding eigenvectors of the training data overlap (kernel) and covariance matrices [2].

SVM have high training performance and low generalization error which pointed out the potential problems of SVMs when the training set is noisy and imbalanced. The SVM is not that much scalable on large data sets because it take time for multiple scanning of data sets hence it is too expensive to perform. To overcome this problem, Clustering-Based SVM (CB-SVM) comes into picture for scalability and reliability of SVM classification [6]. Clustering-Based SVM (CB-SVM) is the SVM technique that is designed for handling large data sets which applies on hierarchical micro-clustering algorithm that scans the entire data set only once to provide the high quality of samples. CB-SVM is scalable if and only if the efficiency of training maximizing the performance of SVMs.

## III. COMPARATIVE STUDY

In this we have done comparative study of Decision Tree and Support Vector Machine, supervised classification techniques, based on predictive accuracy, fitting speed, prediction speed, memory usage and area under curve is shown in Table 1. Further Table 2 shows advantages, limitations and applications of both techniques.

Table1: Comparison between DT and SVM

| Model | DT | SVM |
|---|---|---|
| **Predictive Accuracy** | Low | High |
| **Fitting Speed** | Fast | Medium |
| **Prediction Speed** | Fast | * |
| **Memory Usage** | Low | * |
| **Easy to Interpret** | Yes | * |
| **Handles Categorical Predictors** | Yes | No |
| **Area Under the Curve (AUC)** | More | Less |

Table 2: Advantages and limitations of classification techniques

| Model | DT | SVM |
|---|---|---|
| **Advantages** | • Easy to understand<br>• Easy to generate<br>• Reduce problem capacity | • Ability to learn dimensionality of the feature space. |
| **Limitations** | • Required separate test set<br>• Training time is so expensive<br>• Does not handle continuous variable<br>• Suffer from Overfitting | • Kernel selection<br>• Parameter tuning |
| **Application** | • Text Categorization<br>• Image Classification | • Handwritten Recognition<br>• Text Categorization<br>• Image Classification |

**\*** SVM prediction speed and memory usage are good if there are few support vectors, but can be poor if there are many support vectors. When we use a kernel function, it can be difficult to interpret how SVM classifies data; through the default linear scheme is easy to interpret.

## IV. CONCLUSION

In this report, we saw the different supervised classification techniques on the era of Big Data. Both techniques is better suited than the other for different applications. We also stated a table showing the advantages and disadvantages of the different classification techniques. These techniques can be used to organize all kinds of user needs. Each technique has a different accuracy, speed and predictors. The study indicates that the classification accuracy of SVM algorithm was better than DT algorithm which also gives better classification datasets than DT algorithm.